\documentclass[conference]{IEEEtran}
\IEEEoverridecommandlockouts

\usepackage{cite}
\usepackage{amsmath,amssymb,amsfonts}
\usepackage{algorithm}
\usepackage{algpseudocode}
\usepackage{textcomp}
\usepackage{xcolor}
\usepackage{soul}
\usepackage{graphicx} 
\usepackage{comment}
\usepackage{pdfpages}
\usepackage{xurl}
\usepackage{etoolbox}
\usepackage{hyperref}

\newbool{author_sep}
\boolfalse{author_sep}

\makeatletter
\newcommand{\linebreakand}{
  \end{@IEEEauthorhalign}
  \hfill\mbox{}\par
  \mbox{}\hfill\begin{@IEEEauthorhalign}
}
\makeatother

\def\BibTeX{{\rm B\kern-.05em{\sc i\kern-.025em b}\kern-.08em
    T\kern-.1667em\lower.7ex\hbox{E}\kern-.125emX}}

\title{Multivariate Data Augmentation for Predictive Maintenance using Diffusion\\
\thanks{This material is based upon work supported by the Engineering Research and Development Center - Information Technology Laboratory (ERDC-ITL) under Contract No. W912HZ23C0013. Any opinions, findings and conclusions or recommendations expressed in this material are those of the author(s) and do not necessarily reflect the views of the ERDC-ITL.}
}
\ifbool{author_sep}
{
}
{
    \author{
        \IEEEauthorblockN{
            Andrew Thompson\IEEEauthorrefmark{1}\IEEEauthorrefmark{3},
            Alexander Sommers\IEEEauthorrefmark{1}\IEEEauthorrefmark{3},
            Alicia Russell-Gilbert\IEEEauthorrefmark{1},\\
            Logan Cummins\IEEEauthorrefmark{1}, 
            Sudip Mittal\IEEEauthorrefmark{1},
            Shahram Rahimi\IEEEauthorrefmark{1},\\
            Maria Seale\IEEEauthorrefmark{2},
            Joseph Jaboure\IEEEauthorrefmark{2},
            Thomas Arnold\IEEEauthorrefmark{2},
            Joshua Church\IEEEauthorrefmark{2}
        }
        \IEEEauthorblockA{
            \IEEEauthorrefmark{1}
            \textit{Computer Science \& Engineering} \\
            \textit{Mississippi State University}
            \\\{agt158, ams1988, ar2836, nlc123\}@msstate.edu, \{mittal, rahimi\}@cse.msstate.edu
        }
        \IEEEauthorblockA{
            \IEEEauthorrefmark{2}
            \textit{Engineer Research and Development Center} \\
            \emph{Department of Defence}
            \\\{maria.a.seale, joseph.e.jabour, thomas.l.arnold, joshua.q.church\}@erdc.dren.mil
        }
        \IEEEauthorrefmark{3} corresponding author
    }
}

\begin{document}

\maketitle


\begin{abstract}
    Predictive maintenance has been used to optimize system repairs in the industrial, medical, and financial domains. This technique relies on the consistent ability to detect and predict anomalies in critical systems. AI models have been trained to detect system faults, improving predictive maintenance efficiency. Typically there is a lack of fault data to train these models, due to organizations working to keep fault occurrences and down time to a minimum. For newly installed systems, no fault data exists since they have yet to fail. By using diffusion models for synthetic data generation, the complex training datasets for these predictive models can be supplemented with high level synthetic fault data to improve their performance in anomaly detection. By learning the relationship between healthy and faulty data in similar systems, a diffusion model can attempt to apply that relationship to healthy data of a newly installed system that has no fault data. The diffusion model would then be able to generate useful fault data for the new system, and enable predictive models to be trained for predictive maintenance. The following paper demonstrates a system for generating useful, multivariate synthetic data for predictive maintenance, and how it can be applied to systems that have yet to fail.
\end{abstract}




\section{Introduction}\label{sect::intro}
We rely daily on the assumption that engineered systems will function as expected. However, these systems degrade with use, and eventually fail, wasting valuable time and resources. The solution is maintenance, which raises its own complications. Preventive (early) maintenance can waste useful resources while reactive (late) maintenance can allow catastrophic breakdowns. Predictive maintenance (PdM) aims to solve this problem, predicting breakdowns based off of the current condition of the system. This approach ensures maintenance is performed neither too early nor too late, optimizing resource use and minimizing downtime. Despite its benefits, PdM is complicated by the sparsity of breakdown data relative to functional data \cite{D3A-TS}. This data imbalance exacerbates another difficulty in deploying machine-learning for dynamic systems: concept drift.

Concept drift occurs when the distribution of data shifts during or after the training of an adaptive system. Concept drift can cause a model to become obsolete; the function it has learned no longer reflects the function implicit in the data \cite{concept_drift_survey}. In the context of PdM, concept drift occurs frequently. For example, suppose there exists a set of $n$ jet-engines, $\mathbf{S} = \{ s_{1}, ..., s_{n}\}$, all identical at $t = 0$. These jet engines will be subjected to various stresses as they are used on different planes, for different flights, flown by different pilots, under varying conditions. Consequently, the condition of each engine will not only diverge from that of other engines but also deviate from its original state over time. These varied conditions lead to what is known as concept drift.

Addressing concept drift poses a challenge for PdM, especially when fault data is scarce. The present work explores the use of generative AI to combat concept drift. Exploring the assumption of similarity of degradation between similar systems, generative AI is used to create synthetic signals indicative of degeneration in otherwise healthy systems to preemptively train a PdM model. 

However, data complexity presents a hurdle for generating synthetic data. As sensors increase in complexity, so does the data they record \cite{IECMA2022-12901}. Sensor improvements have caused data sampling rates to increase, meaning that more data points are recorded during time windows. As different types of sensors are combined and improved, multivariate data samples are becoming more common. The increase in sample length and number of variables requires synthetic generative methods to improve their capabilities. They must have the ability to capture long term dependencies, and effectively generative  data of a higher complexity. Previous work in this area has not targeted complex PdM tasks, limiting the application of these methods. Modern PdM tasks have become increasingly challenging; therefore, this paper employs cutting edge generative models to assist in training PdM models.

The contributions of this work are:
\begin{itemize}
    \item Designing and implementing a method to showcase the effectiveness of diffusion models for complex multivariate predictive maintenance tasks.
    \item Testing DSAT-ECG on the PRONOSTIA dataset, widely used in PdM. 
    \item A method for combating concept drift by inferring likely fault signals for still healthy systems.
\end{itemize}

The remaining sections of this paper are as follows.
Section \ref{sect::prior_art} discusses the relevant background information. Section \ref{sect::core} explains the methods used. Section \ref{sect::results} lays out the results of our testing. Section \ref{sect::disc} analyses the results. Lastly, Section \ref{sect::conc} concludes the paper with a discussion of future works and final remarks.



\section{Background and Prior Art}\label{sect::prior_art}
The lack of fault data has caused many PdM tasks to have imbalanced datasets, making it difficult to train effective classifiers. Many methods have been devised to try and mitigate imbalanced data, with varying results. Oversampling methods increase the number of instances of the minority class while undersampling methods decrease the number of instances of the majority class. Several of these sampling methods, including SMOTE, ADASYN, MTDF, MWMOTE, TRkNN and ICOTE have shown considerable success in addressing data imbalances \cite{Sampling}, but are not suitable for this target use case. Overfitting, and the inability to extrapolate beyond existing data, limit their effectiveness in PdM cases where the data of systems that have yet to fail must be inferred. Despite these limitations, they hint at a more promising method of handling imbalanced data: synthetic data generation.

Synthetic generation is the process of generating new, artificial samples using known data samples. A simple generation method is showcased in SMOTE (Synthetic Minority Over-sampling Technique) \cite{SMOTE}. SMOTE creates synthetic data by selecting minority class instances, identifying their nearest neighbors in the feature space, and creating new samples by randomly selecting points along the line segments connecting the instance to its neighbors. This only works for certain datasets, as these types of sampling methods rely on linear properties and distinct class features. Recent studies into PdM have used Generative Adversarial Networks (GANs) to generate high-quality synthetic data \cite{Anchor_for_GANS_in_PdM_data_aug, GPDH1, GPDH2, GPDH3, GPDH4, GPDH5, GPDH6, GPDH7}. These generative methods have been used to preserve data privacy by providing shareable synthetic counterparts to confidential data \cite{rashid}.  GANs frequently encounter the problem of mode collapse, where the model fails to capture the full diversity of the training data. Diffusion models can also generate high-quality samples and do not suffer from mode collapse, challenging the dominance of GANs. \cite{Qian_2024_CVPR} The compatibility of diffusion models and time series data makes them further applicable to PdM, since most sensors report time series data.

Diffusion models work by incrementally adding Gaussian noise to data samples throughout a forward process. They learn to recover the original data by removing the added noise through a reversal process. To remove noise, a neural network is trained to predict noise added to the sample. That predicted noise is then removed from the data sample during the reversal process. Synthetic samples can then be generated by inputting random noise samples into the model, and having the reversal process remove noise \cite{DiffusionSurvey}. Some modern diffusion models incorporate conditioned diffusion. With this technique, data generation can be guided using passed in variables. Their latent representation is usually concatenated with the input data and passed into the neural network.

Diffusion models have shown noticeable improvement in recent years, starting with DiffWave in 2021 \cite{DiffWave}. DiffWave was inspired by WaveNet \cite{WaveNet}, a model that used the ResNet architecture \cite{ResNet} to generate audio samples. DiffWave was the first to implement the diffusion process into the ResNet architecture. CSDI \cite{CSDI} followed, adding attention mechanisms to more powerfully model time-series features. Once S4 \cite{S4} was introduced, SSSD\textsuperscript{S4} \cite{SSSD_S4} and CSDI\textsuperscript{S4} took advantage of state-space model's ability to model time-series data. SSSD-ECG \cite{SSSD_ECG} was produced by the same authors, applying their model to ECG data. DSAT-ECG \cite{DSAT-ECG} applied the new SPADE block, built upon the TNN and S4 modules. This model will be used in our primary work, thanks to the code generously provided by the authors of DSAT-ECG. Time Weaver \cite{TimeWeaver} improved the conditioning pipeline, but does not incorporate the SPADE block. 

This recent progress makes diffusion a promising tool, with the D3A-TS model \cite{D3A-TS} already applying it to data augmentation. D3A-TS is a method that uses meta-attributes to condition the denoising model. This method enhances a diffusion model's ability to generate synthetic data. D3A-TS was conducted using 6 time series datasets, with a few being PdM datasets. Mart{\'i}n et al. were able to successfully apply a diffusion model to augment their datasets. While they included PdM datasets in their testing, their work was not focused on conducting PdM. They did not attempt to use diffusion to anticipate unavailable fault data. This is crucial due to concept drift affecting many maintenance systems \cite{ConceptDrift}. Additionally, we are attempting to generate data of a higher complexity, both multivariate and with many more time steps in each sample. Our work explores this gap in the literature, and utilizes recent advancements in diffusion models, including the use of state space models in order to generate extended and detailed fault data that can be useful for PdM.



\section{Methodology}\label{sect::core}
This project sought to test the hypothesis that diffusion generated synthetic data can successfully supplement maintenance datasets to improve predictive models. If fault data exists for a subset of systems, are we able to conditionally generate synthetic data that can be used to supplement the target dataset?

Fault data may be scarce or unavailable due to malfunctioning systems tending to shut down and quickly receive repairs. The system might also be new enough to never have reached a failure point, making fault data nonexistent. Concept drift can cause fault data to vary significantly between systems using similar devices. A secondary goal of this project was to determine if a diffusion model can generate useful synthetic fault data for a system that has yet to fail by using run-to-failure data from a similar system along with healthy data from the target system.

\subsection {Model}
\begin{figure}
    \centering
    \includegraphics[width=0.95\linewidth]{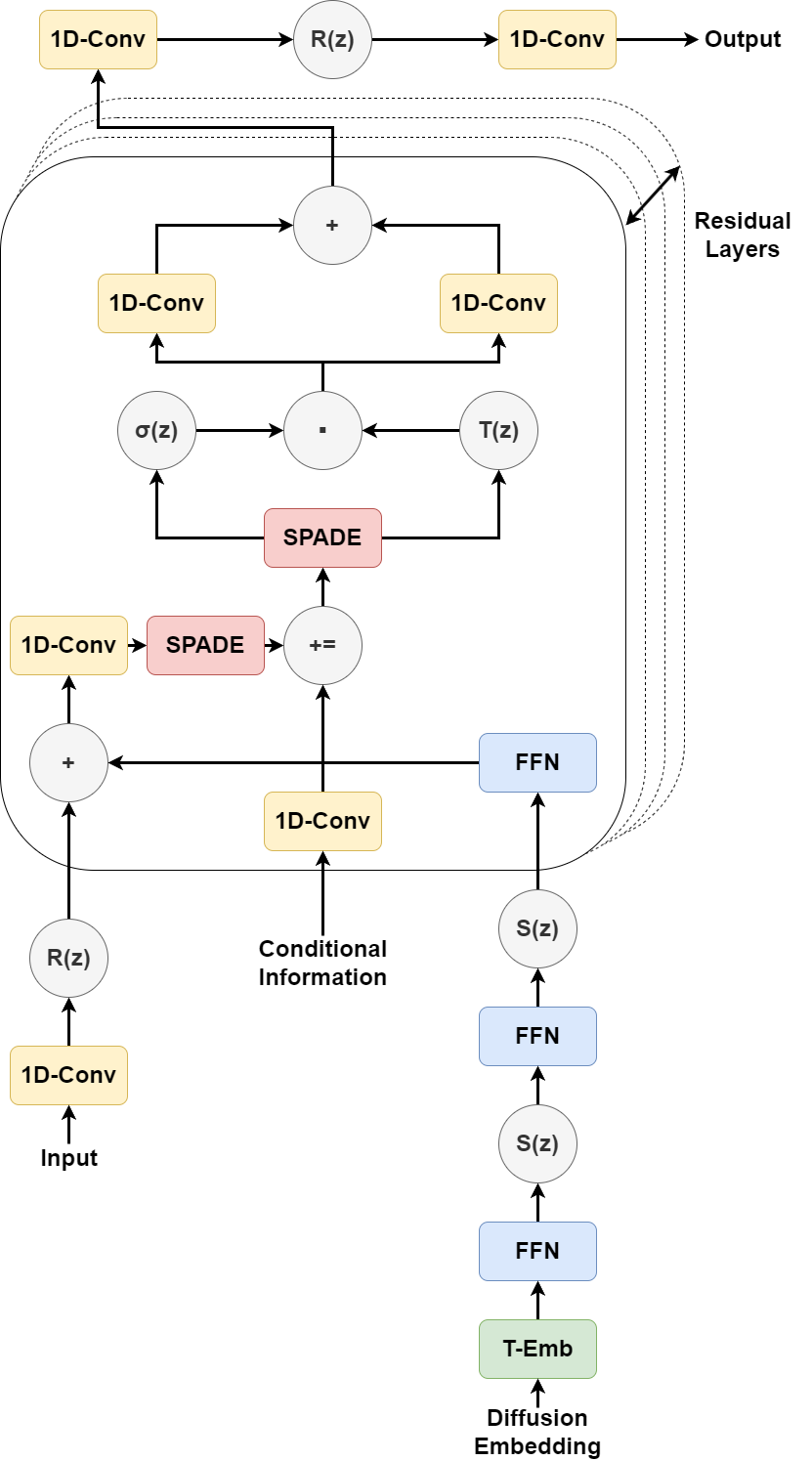}
    \caption{DSAT-ECG model architecture. Figure inspired by \cite{DSAT-ECG}. FFN stands for Feed-Forward Network, and T-Emb stands for Time Embedding. DSAT-ECG is able to apply conditional information to guide the denoising process in generating certain types of data samples. The SPADE blocks are able to help capture both local and long term dependencies.}
    \label{fig:DSAT-ECG}
\end{figure}

To generate synthetic data, this project used the DSAT-ECG model \cite{DSAT-ECG} seen in Fig. \ref{fig:DSAT-ECG}. The lineage of DSAT-ECG can be found in Section \ref{sect::prior_art}. This model was chosen due to its ability to create high quality synthetic data and capture long term dependencies. Time Weaver was not used due to a lack of paired meta conditions in the target dataset along with its exclusion of SPADE blocks.

\subsection{Dataset}
This project used the PRONOSTIA dataset \cite{PRONOSTIA} to evaluate the possibility of effectively using diffusion for PdM. PRONOSTIA contains runs-to-failure created using the PRONOSTIA platform. This platform subjected ball-bearings to a variety of destructive conditions. A vibration sensor was placed on both the x-axis and y-axis to record data. Each use of a bearing was halted when the vibration amplitude exceeded 20 g. In order to look at predictive maintenance scenarios, the dataset must be partitioned so that the fault data of targeted bearings can be predicted. Any partial runs were not included in our testing. Only 6 runs go to failure, those runs being 3 unique bearing conditions stressed for 2 different runs.

\begin{figure}[htbp]
    \centering
    \includegraphics[width=0.45\textwidth]{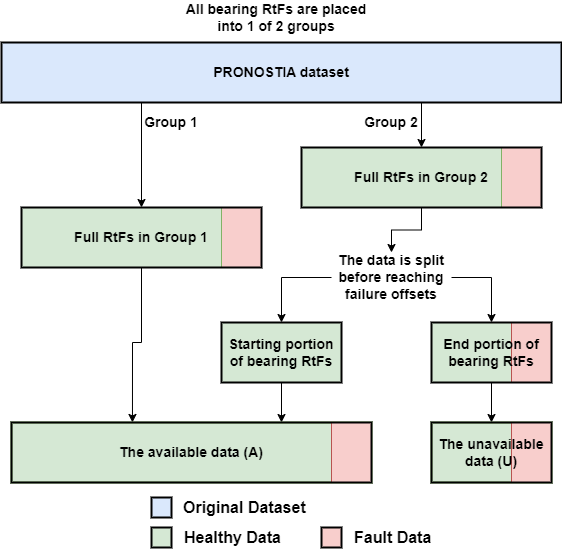}
    \caption{The dataset runs-to-failure (RtF's) are divided into two disjoint groups. Dataset \textbf{A} comprises complete RtF data of the Group 1 bearings. Additionally, it includes the initial operational data for the Group 2 bearings prior to any faults. Dataset \textbf{U} encompasses the remaining operational data from each RtF for the Group 2 bearings.}
    \label{fig:partScheme}
\end{figure}

\subsection{Partitioning Scheme}
    Under the assumption that the dataset consists of $n$ complete run-to-failure (RtF) records with $d$ features over time, let:
    \begin{itemize}
        \item $o$ be the offset backwards in time from the end-of-life (EoL). Known as a failure horizon \cite{FailHorizon}. For this study, the EoL begins at the first data point that is 3 or more standard deviations away from the mean of the sample.
        \item $k$ be the number of complete RtFs in the ``available" partition $A$. These RtFs are ``complete", and others are ``incomplete".
        \item $\gamma$ modifies the amount of healthy data from the $n-k$ RtFs that can be in $A$.
        \item $U$, the ``unavailable" partition, be all data not in $A$.
    \end{itemize}

     Each data sample is a collection of 2560 data points that include both horizontal and vertical acceleration features, making the size of each sample (2560, 2).
    $o$ is a number of time-steps, and each time-step within that window backwards from the EoL of an RtF is defined as anomalous data. $\gamma$ is a percentage of each unavailable bearing that is included in $A$. For example, a $\gamma$ = 0.2 includes the front 20\% of each unavailable bearing in $A$. If fault data is found before reaching $\gamma$ (20\%), the cutoff for $A$ ends right before the first fault point. This enables us to train our model to attempt to generate fault data of a type it has not seen before. For our data supplementation that utilizes all available fault data, the entire dataset was used instead.

\subsection {Evaluation Methods}
To evaluate the effectiveness of using diffusion for PdM, the model must first prove able to generate high quality samples of the chosen dataset (PRONOSTIA). We evaluated this capacity using the TSGBench benchmark \cite{TSG_benchmark}, a widely used testing criteria for data generation models. This way our metrics can be compared to other works in the field. TSGBench uses 12 metrics for testing: \begin{itemize}
    \item Discriminative Score (DS)
    \item Predictive Score (PS)
    \item Contextual-FID (C-FID)
    \item Marginal Distribution Distance (MDD)
    \item Auto Correlation Distance (ACD)
    \item Skewness Difference (SD)
    \item Kurtosis Difference (KD)
    \item Euclidean Distance (ED)
    \item Dynamic Time Warping (DTW)
    \item Training Time (Time)
    \item t-SNE plots
    \item Distribution Plots
    \end{itemize}
 For the purpose of this study, we excluded training time and the distribution plots. Training time measures how long a test takes to run to completion, not how long the diffusion model trained, which we have included. The distribution plot was deemed redundant information, which was why it was excluded. In addition to TSGBench, we sought to test the effectiveness of our synthetic data. If multiple predictive models trained using our synthetic data outperform those trained without it, we can infer our data is useful. To measure model success, we recorded precision, recall, and f1-score for each class (healthy and faulty).

To test the effectiveness of our data, the following were used:
\begin{itemize}
    \item A generative model $M_g$ based on DSAT-ECG \cite{DSAT-ECG}.
    \item A PdM-task predictive model $M_p$, \cite{Cummins2024ExplainableAD}
    \item An RtF dataset (PRONOSTIA). 
\end{itemize}
\vspace{0.3cm}
The procedure was as follows:
\begin{enumerate}
    \item Partition the dataset into $A$ (available) and $U$ (unavailable) using $\gamma$ = 0.3.
    \item The generative model $M_{g}$ is trained on all of $A$ and is then used to create $S$, a synthetic counterpart to $U$. This requires the use of a conditioning scheme such that the metadata of $A$ is used to guide the generation of $S$. The conditional signal consists of bearing ID and a binary value for health (0 being healthy and 1 being faulty). We assume some healthy data in $A$ for every bearing ($\gamma \neq 0$). This tests our hypothesis. If $M_{g}$ has been trained to generate healthy and unhealthy data using $A$, can it generate a useful approximation of what degradation signals will look like for a bearing \emph{incompletely} in $A$, a bearing in $U$?
    \item Two instances of $M_{p}$ are trained, $M_{p}'$ on $A$, $M_{p}''$ on $A+S$. Both are tested on $U$. The results are assessed by looking at the precision, recall, and f1-scores of each model. If $M_{p}''$ performs better than $M_{p}'$, then our hypothesis would be supported.  
\end{enumerate}

\begin{figure}[htbp]
    \centering
    \includegraphics[width=0.45\textwidth]{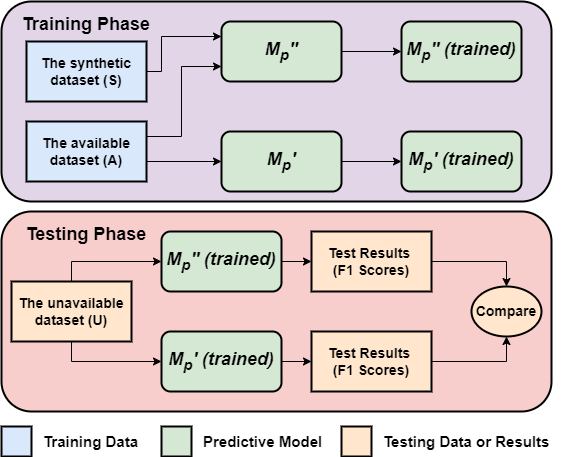}
    \caption{To test the quality of synthetic data, 2 instances of our predictive model ($M_{p}$) were trained. One model ($M_{p}'$) is trained on just the original dataset \textbf{A}. The other model ($M_{p}''$) is trained using both the original dataset \textbf{A} and the synthetic dataset \textbf{S}. If ($M_{p}''$) performs better, the synthetic data can be considered useful for PdM.} 
    \label{fig:testingScheme}
\end{figure}

\begin{figure*}[tb]
    \centering
    \includegraphics[width=0.95\linewidth]{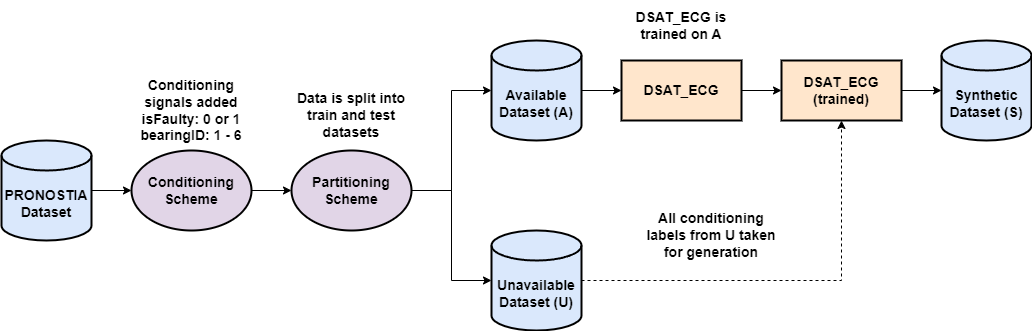}
    \caption{Generation scheme used for DSAT-ECG generation of unavailable data. Details for the partitioning scheme can be found in Figure \ref{fig:partScheme}. Conditional labels are taken from U for generation, as TSGBench works best when the real and synthetic data samples are equal in number and type.}
    \label{fig:genScheme}
\end{figure*}

\subsection{The Conditioning Scheme}

Conditional generation requires a descriptive conditional signal. Since the model is involved with levels of degradation, the conditional signal of $M_{g}$ and the output value of $M_{p}$ are related, but are not necessarily the same.

$M_{g}$ outputs and $M_{p}$ receive an input, a window $w \in \mathbf{R}^{d \times l}$ where $l$ is the number of time-steps and $d$ is the number of covariates. Thus the conditioning signal, hereafter the ``con-signal", should concern a window over time, not a data point in time. In simulation of the target use case, it is reasonable to assume a partition such that total historical data for all failed systems (entire RtFs) are available, and that the remaining useful life (RUL) of any system that has yet to fail \emph{can not} be known. Thus, the con-signal can not communicate a \% lifetime, since that can only be known retrospectively. Since each window generated is of length $l$, the size of the generated window is not part of the con-signal. Even over short periods, the system (a bearing in PRONOSTIA) changes, and this becomes more true as $l$ increases. System ID is a catch-all for the unknown idiosyncrasies of any given system-instance, and is therefore included in the con-signal. In addition to system ID (Bearing ID for PRONOSTIA), a binary label for fault was applied to each label (0 for healthy and 1 for faulty). If any data point in a window is within the failure horizon, the entire sample is labeled faulty. This only impacted 1 sample in each RtF, as each remaining sample was completely within the failure horizon. The con-signal therefore consists of 2 attributes, Bearing-ID and isFaulty. The Bearing-IDs were one-hot-encoded, with the isFaulty label concatenated onto the end of each con-signal. It is feasible, and potentially advantageous, to include more conditional variables; however, our focus here was on creating a more general conditioning scheme.

Central to testing the hypothesis of this work was examining if such a con-signal scheme, applied to a diffusion model that has been trained on the available data, can be used to supplement the unavailable data. When a complete RtF is part of the available dataset, the model will possess fault data for that bearing ID. One of the questions this paper sought to answer is whether or not the model can determine what makes a sample "faulty" and apply that knowledge to pure healthy data in order to create unavailable fault data.

\subsection{The Generative Scheme}

DSAT-ECG was used in an attempt to generate fault data for bearings whose failure data was not used in training the model. $S$ is an attempt to replicate $U$ such that training $M_p$ on $A \cup S$ results in better performance, tested on $U$, than training on $A$ alone. To rigorously define our approach, $S$ had to be specified, including such features as the number of generated windows, the proportion of them that are healthy, and the length of those windows. A generative scheme must bound and specify $S$.

For our purposes, we used all of the labels in $U$ as our con-signals, with $b$ specifying the bearing ID in the PRONOSTIA dataset, and $f$ determining if a given synthetic window is faulty. This way, the ratio of healthy to faulty samples stays consistent between $S$ and $U$, improving TSGBench's effectiveness as a metric for comparison. The size of each generative window was set to 2560 to mimic the size of the real data. Two outputs were defined: horizontal and vertical acceleration. An outline of the generative process can be found in Figure \ref{fig:genScheme}. DSAT-ECG was also used to generate supplementary fault data. For this task, all available fault data was utilized, meaning that both $U$ and $A$ were used to train the model.



\section{Results}\label{sect::results}
\begin{figure}[htbp]
    \centering
    \includegraphics[width=1.0\linewidth]{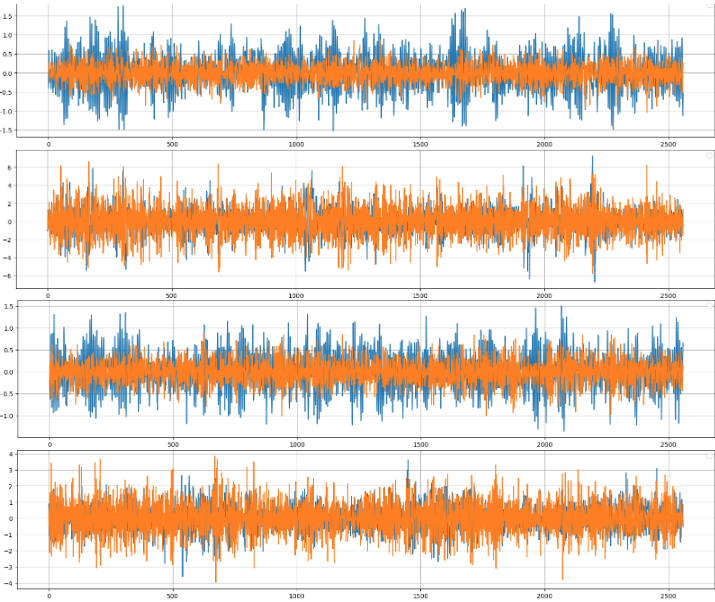}
    \caption{Samples of horizontal (orange) and vertical (blue) bearing data. From top to bottom: real healthy sample, real faulty sample, synthetic healthy sample, synthetic faulty sample. Values range from (-1.5, 1.5), (-6, 6), (-1, 1.5), and (-4, 4) respectively.}
    \label{fig:samples}
\end{figure}

\begin{figure}[htbp]
    \centering
    \includegraphics[width=1.0\linewidth]{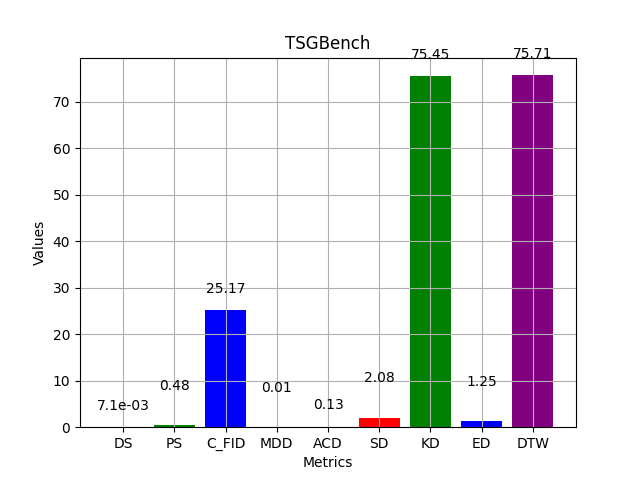}
    \caption{This graph depicts the results of metrics comparing our synthetic data against real data using TSGBench. This model was trained using all available data.}
    \label{fig:TSGBenchResults}
\end{figure}

\begin{figure}[htbp]
    \centering
    \includegraphics[width=0.8\linewidth]{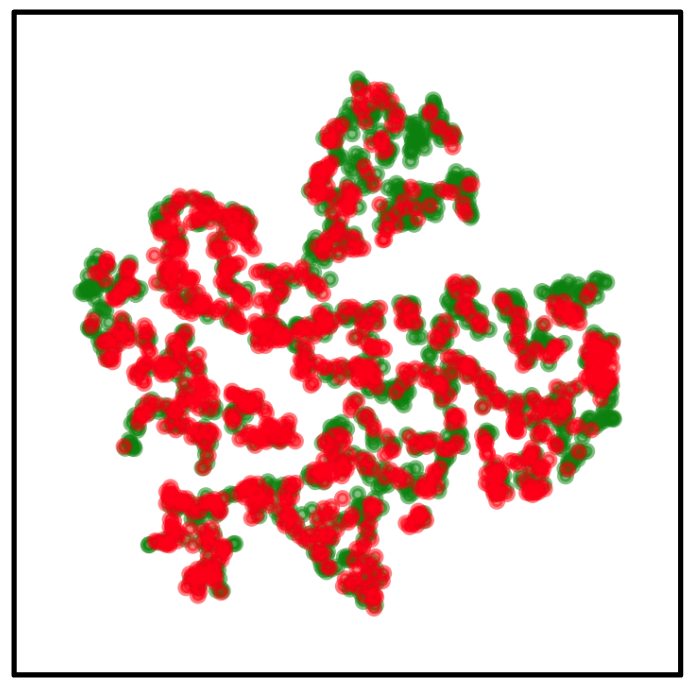}
    \caption{t-SNE graph of real data (green) and synthetic data (red) each representing their dataset projected into a lower dimensional space.}
    \label{fig:t-SNE}
\end{figure}

\begin{figure}[htbp]
    \centering
    \includegraphics[width=1.0\linewidth]{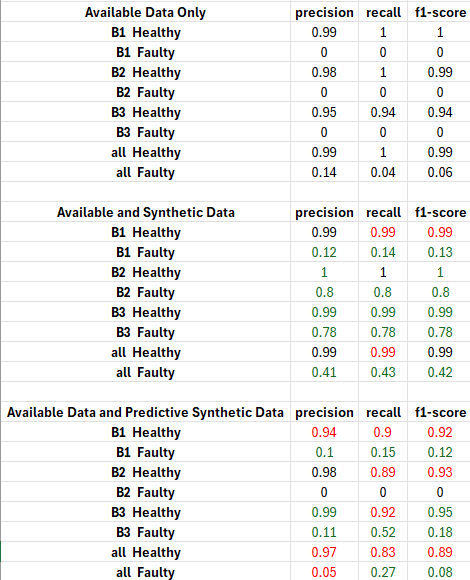}
    \caption{ The top section of predictive models are the default model without any synthetic data. The middle models include synthetic data generated by a DSAT-ECG model trained on the entire dataset. The bottom models include synthetic data generated by a DSAT-ECG model that was trained without any of the relevant fault data. The bottom 2 models show how our synthetic data affected model training.}
    \label{fig:finalResults}
\end{figure}

The following figures showcase the results of our generative method. Figure \ref{fig:samples} shows examples of healthy and faulty data from the actual dataset, along with synthetic samples generated by DSAT-ECG. Figure \ref{fig:TSGBenchResults} gives the results of TSGBench. Discriminative Score (DS) shows how well a predictive model was able to correctly identify the difference between real data and our synthetic data. A score of 0 is equivalent to an accuracy of 50\%, which is the same as randomly guessing. Predictive Score (PS) showcases how well a post-hoc time series prediction model trained on synthetic data compares to one trained on the real data. Contextual FID (C-FID) quantifies how well the data conforms to local context. Marginal Distribution Difference (MDD) measures how closely the distributions of the original and generated data series align. AutoCorrelation Difference (ACD) shows how well dependencies from the original dataset are maintained in the synthetic data samples. Skewness Difference (SD) measures the difference between asymmetries of the distributions. Kurtosis Difference (KD) compares the outliers of each dataset. Euclidean Distance (ED) measures the similarity between real and synthetic data samples by calculating the straight-line distance between corresponding values across all samples. It does this by iterating through each sample pair and comparing their individual data points. Dynamic Time Warping (DTW) shows the temporal alignment between both datasets. Figure \ref{fig:t-SNE} visualizes both datasets in a lower dimensional space. While these evaluation metrics can be useful in highlighting certain characteristics, none of them are able to measure the usefulness of our synthetic data, which is why we included predictive models (Figure \ref{fig:finalResults}). Three batches of predictive models were trained. Each batch consisted of 4 models. The first batch only used the real dataset. This was the baseline for our test. The second batch trained using both the real dataset, and a supplement of synthetic data. The synthetic data used for this batch was created using all available data for training. For systems that have yet to fail, this method would be invalid, as no fault data would be available to assist in training the diffusion model. For this reason, the third batch was created. The predictive models for each type of bearing run were trained using all available real data \emph{except} their own fault data, along with synthetic data from a diffusion model trained using the method found in Figure \ref{fig:partScheme}. This enables the predictive models to be trained without requiring \emph{any} fault data from their system.

\begin{figure}[!htbp]
    \centering
    \includegraphics[width=.95\linewidth]{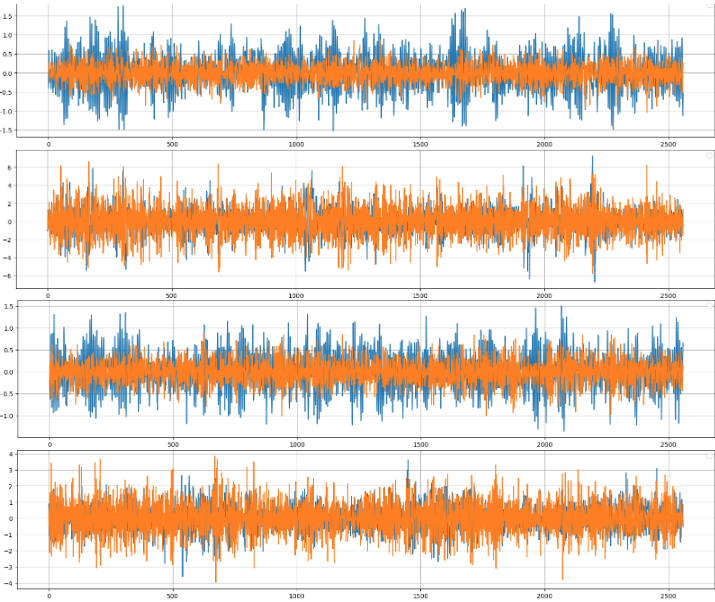}
    \caption{Actual dataset compared to synthetic samples generated from DSAT-ECG. From top to bottom: real healthy sample, real fault sample, synthetic healthy sample, synthetic fault sample. All samples are from bearing 2.}
    \label{fig:dataShow}
\end{figure}

\begin{figure}[htbp]
    \centering
    \includegraphics[width=1.0\linewidth]{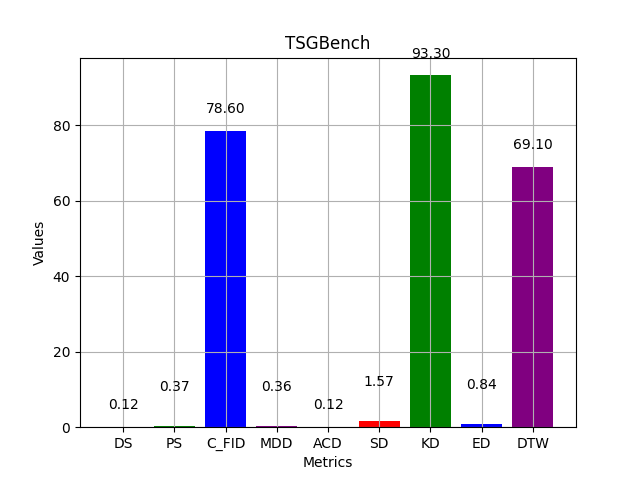}
    \caption{This graph depicts the results of metrics comparing our predictive synthetic data against real data using TSGBench. This model trained with no fault data from its target bearing.}
    \label{fig:TSGBenchResults2}
\end{figure}

\section{Discussion}\label{sect::disc}
After performing TSGBench on our synthetic data, we recorded the t-SNE graph and saved 7 of the above mentioned benchmark metrics, with lower numbers being better. For DS and PS, the range of values is limited to (0,1). Our results (Figure \ref{fig:TSGBenchResults}) show overwhelming success in almost every category, showcasing the effectiveness of diffusion models in synthetic generation. There were 2 categories with poor results: Dynamic Time Warping (DTW) and Kurtosis Difference (KD). Poor DTW results were somewhat expected due to the complex temporal dependencies in the dataset along with the lack of detail in the conditioning scheme for such a metric. Our goal was not to generate an entire RtF, but to instead generate individual sample windows of high quality synthetic data. DTW is of relative unimportance for our use case, as our predictive model does not rely on temporal characteristics between individual samples. If our model was designed to predict the order of a sample, this metric would be crucial. Our subpar KD score resulted from the difficulty our model had in generating synthetic fault data of equal amplitude to the real data (as show in figure \ref{fig:samples}). Sparsely occurring in the real data were major spikes in vibration, exceeding 10-20 times normal operating conditions. These values are either not captured in the synthetic data or are reduced in magnitude, further hurting our KD score. Every other metric was within acceptable ranges \cite{TSG_benchmark}, showcasing the similarity between the real and synthetic data. A positive metric to point out is the model's Discriminatory Score, getting near perfect marks. Overall, the results showcase the ability to supplement predictive maintenance datasets with synthetic data. Our more ambitious goal, to predict unavailable fault data using fault data from similar systems and healthy data from the target system, was not nearly as successful. The predicted fault samples were only marginally different from their healthy counterparts, and suffered poor results in C-FID (78.6), KD (93.3), and DTW (69.1) \ref{fig:TSGBenchResults2}. The model was unable to transfer the faulty/healthy relationship of other bearing runs to its target bearing. Results suggest that this is partially attributed to diminishing strength of the faulty/healthy relationship for healthy data further into the RtFs. 

Regarding the effectiveness of our data, we used \cite{Cummins2024ExplainableAD} to attempt to identify faulty samples. Without fault data from the target bearing, the models were unable to successfully identify faulty samples. Trained with an infusion of synthetic data, the models showed useful improvement. They were able to detect faults at a much higher rate, with improvements reaching 80\% \ref{fig:finalResults}. The complexity and multivariate nature of our dataset made this feat more impressive, and opens the door to more applications of synthetic supplementation. Our synthesized fault samples were also able to improve the performance of the predictive model in certain categories, but overall were not substantially useful. Our findings suggest that the increasing similarity between healthy samples near the end of RtFs and fault samples made it more challenging to accurately learn the characteristics of the fault label. As the RtFs progressed, healthy data became statistically similar to failure data. The model was unable to correctly apply fault characteristics to create unseen types of fault data. However, subtle indications of the model attempting to produce such fault samples
suggest the value of pursuing future work in this area.



\section{Conclusions and Future Work}\label{sect::conc}
Future work for this project involves improving the diffusion model and applying it to more datasets. If the amplitude of sparsely represented data spikes can be noticed by the model it is believed the effectiveness of synthetic data for training could be applied to even more complex datasets. This method is also partially reliant on SMOTE for training the generative model when the minority class is unable to transfer its characteristics over to the diffusion model. Regarding training time, we were able to produce these results with 9 hours of training on a RTX 4090. In our opinion, this is a acceptable time frame for useful application of the model to real life scenarios. While predicting unavailable fault data produced only minor success, it did show promise for this capability to be feasible in systems with similar state relationships or with less complexity. Modifications to improve the transfer learning of DSAT-ECG is believed to be a viable goal. We believe this paper has showcased the recent advancements in diffusion, and how it can now be applied to generate complex maintenance data.



\bibliographystyle{ieeetr} 
\bibliography{ref} 


\end{document}